\documentclass{article}

\usepackage[preprint]{neurips_2025}

\usepackage[utf8]{inputenc} % allow utf-8 input
\usepackage[T1]{fontenc}    % use 8-bit T1 fonts
\usepackage{hyperref}       % hyperlinks
\usepackage{url}            % simple URL typesetting
\usepackage{booktabs}       % professional-quality tables
\usepackage{amsfonts}       % blackboard math symbols
\usepackage{nicefrac}       % compact symbols for 1/2, etc.
\usepackage{microtype}      % microtypography
\usepackage{amsmath}
\usepackage{amssymb}
\usepackage{graphicx}
\usepackage{pifont}  
\usepackage{mathrsfs}
\usepackage[table]{xcolor}

\title{Discovering Pathology Rationale and Token Allocation for Efficient Multimodal Pathology Reasoning}

\author{%
  Zhe Xu\thanks{These authors contributed equally.} \\
  Department of Computer Science Engineering\\
  HKUST, Hong Kong SAR, China \\
  \texttt{csezhexu@ust.hk} \\
  \And
  Cheng Jin\footnotemark[1] \\
  Department of Computer Science Engineering\\
  HKUST, Hong Kong SAR, China \\
  \texttt{cheng.jin@connect.ust.hk} \\
  \AND
  Yihui Wang\\
  Department of Computer Science Engineering\\
  HKUST, Hong Kong SAR, China \\
  \texttt{ywangrm@connect.ust.hk} \\
  \And
  Ziyi Liu \\
  Department of Computer Science Engineering\\
  HKUST, Hong Kong SAR, China \\
  \texttt{cseziyiliu@ust.hk} \\
  \And
  Hao Chen\thanks{Corresponding author} \\
  Department of Computer Science Engineering, \\Department of Chemical and Biological Engineering\\
  HKUST, Hong Kong SAR, China \\
  \texttt{jhc@cse.ust.hk} \\
}

\begin{document}

\maketitle

\begin{abstract}

Multimodal pathological image understanding has garnered widespread interest due to its potential to improve diagnostic accuracy and enable personalized treatment through integrated visual and textual data. However, existing methods exhibit limited reasoning capabilities, which hamper their ability to handle complex diagnostic scenarios. Additionally, the enormous size of pathological images leads to severe computational burdens, further restricting their practical deployment. To address these limitations, we introduce a novel bilateral reinforcement learning framework comprising two synergistic branches. One reinforcement branch enhances the reasoning capability by enabling the model to learn task-specific decision processes, \emph{i.e.}, pathology rationales, directly from labels without explicit reasoning supervision. While the other branch dynamically allocates a tailored number of tokens to different images based on both their visual content and task context, thereby optimizing computational efficiency. We apply our method to various pathological tasks such as visual question answering, cancer subtyping, and lesion detection. Extensive experiments show an average +41.7 absolute performance improvement with 70.3\% lower inference costs over the base models, achieving both reasoning accuracy and computational efficiency. 
\end{abstract}

\section{Introduction}

Pathology ~\cite{he2024foundation,ikezogwo2023quilt} serves as the gold standard in modern medicine, providing critical insights into disease mechanisms, diagnosis, and therapeutic decision-making. 
The digitization of pathological images has revolutionized this field, enabling computational approaches to assist in image analysis, thereby improving diagnostic consistency and efficiency. Early computational pathology models ~\cite{chen2016dcan,uni} primarily focused on single-modal image analysis, employing convolutional neural networks (CNNs) and other deep learning techniques to detect and classify diseases based solely on visual patterns. However, real-world pathological diagnosis is seldom limited to visual assessment alone; it often involves integrating information from diverse sources such as clinical notes and patient history. This multimodal nature has driven the shift from single-modal models to more sophisticated multimodal frameworks ~\cite{huang2023visual,xiang2025vision} capable of integrating diverse data types for comprehensive analysis.

The emergence of multimodal large language models (MLLMs) \citep{tong2024eyes, zhang2024cambrian,liu2023visual,liu2024improved,bai2025qwen2,pathasst,pathchat,quiltllava} has further expanded the capabilities of computational pathology, allowing for joint modeling of visual and textual data. Despite their promise, current multimodal pathological models suffer from two fundamental limitations that hinder their clinical applicability:

\textbf{(1) Lack of reasoning capabilities.}
Pathological diagnosis is inherently a reasoning-intensive process. For instance, distinguishing between subtypes of cancer often involves evaluating cellular morphology, tissue architecture, immunohistochemical profiles, and clinical correlations—a multi-step analytical process that mimics human expert reasoning. However, existing multimodal models in pathology are predominantly trained using standard supervised fine-tuning (SFT) that heavily rely on large amounts of supervised data to enhance model performance. The absence of enough explicit reasoning supervision (\emph{i.e.}, \textbf{pathology rationale}) means that models may learn superficial correlations rather than true diagnostic logic, leading to struggle with complex, real-world diagnostic scenarios that require logical inference, contextual understanding, and hierarchical decision-making.

\textbf{(2) Computational inefficiency due to high-resolution images.}
Pathological images are characterized by exceptionally high spatial resolution, often comprising millions of pixels per sample. Processing such high-resolution images imposes severe computational burdens, including excessive memory consumption, prolonged inference times, and high operational costs. Current approaches typically employ static tokenization strategies, where images are encoded into fixed-size tokens regardless of their content complexity. As a result, computational resources are wasted on simple images and insufficiently allocated to more complex, diagnostically challenging cases. These inefficiencies undermine the scalability of computational pathology, particularly in resource-constrained clinical settings.

\begin{figure}
    \centering
    \includegraphics[width=\linewidth]{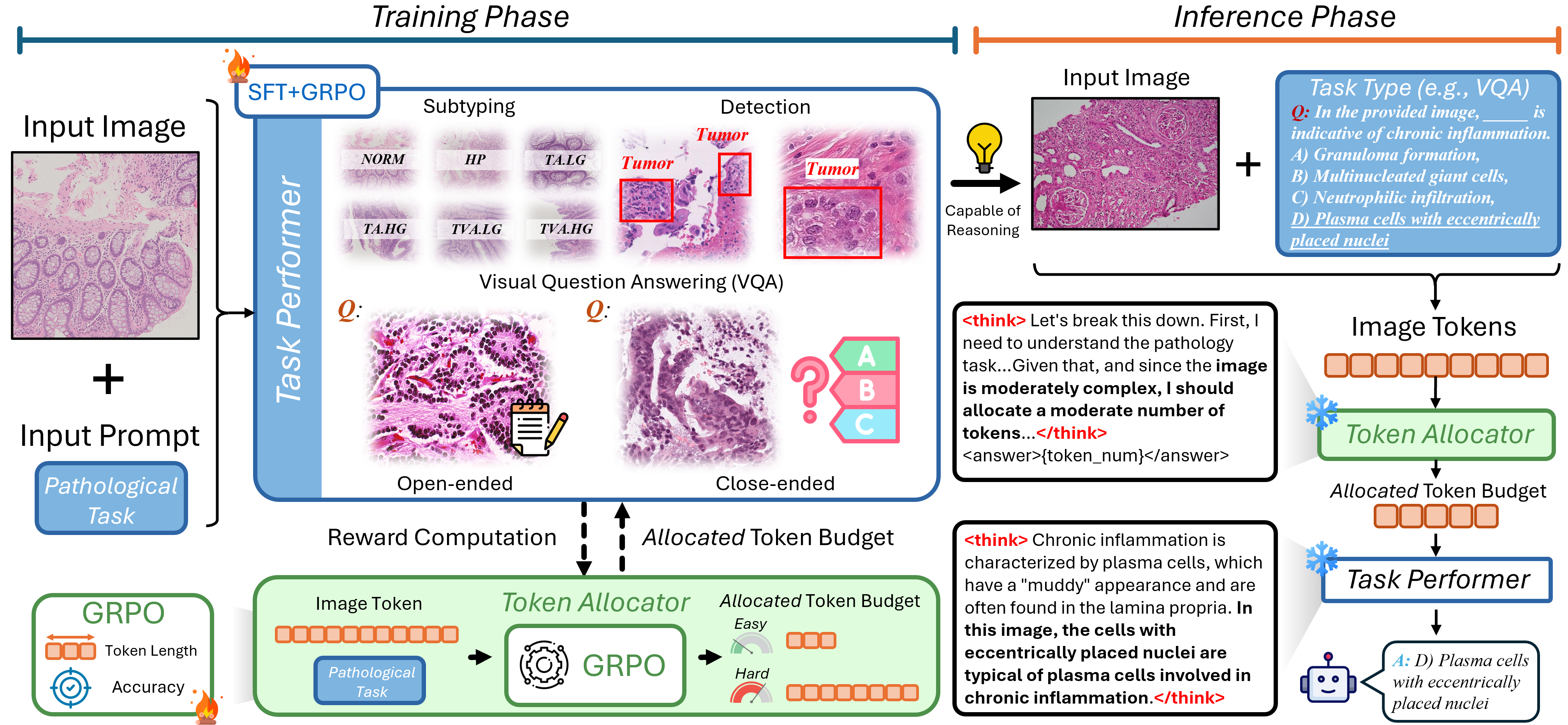}
    \caption{Overview of our framework’s ability to discover underlying pathology rationale and optimize token allocation, enabling efficient and accurate multimodal reasoning for tasks such as VQA, cancer subtyping, and lesion detection.}
    \label{fig:workflow}
\end{figure}

To address these challenges, we introduce a novel bilateral reinforcement learning (RL) framework that simultaneously enhances diagnostic reasoning and optimizes computational efficiency. As shown in Fig. \ref{fig:workflow}, our approach consists of two synergistic branches, \emph{i.e.}, task performer branch and token allocator branch.
Unlike traditional supervised learning that directly optimizes for task output, the task performer employs reinforcement learning to train the model to emulate the reasoning process of pathologists.
By framing diagnostic decision-making as a sequential policy optimization problem, the model learns to generate pathology rationales without requiring explicit supervision. This approach mimics the human diagnostic workflow, where pathologists iteratively gather evidence, weigh competing hypotheses, and refine their conclusions.
To tackle computational inefficiency, we introduce an adaptive tokenization method that dynamically allocates computational resources based on image content and clinical relevance. Instead of processing all images uniformly, the token allocator learns to assign more tokens to diagnostically complex images while reducing redundancy in less informative images. This strategy is guided by both visual saliency and contextual clinical information, ensuring that computational effort aligns with diagnostic importance.

We evaluate our framework on a diverse set of pathological tasks, including visual question answering (VQA), cancer subtyping, and lesion detection. Our experimental results show substantial advancements beyond current approaches, achieving an average 41.7-point improvement in task performance over baseline models on complex pathological assessments. Moreover, the dynamic token allocation mechanism reduces inference costs by 70.3\%, making the framework feasible for clinical deployment.

In summary, our contributions are threefold:
\begin{itemize}
    \item  We introduce the first reinforcement learning paradigm that enables models to learn pathology rationale implicitly from labels, eliminating the need for costly intermediate supervision and improving interpretability.
    \item  We propose a novel token allocation method that dynamically adjusts computational resources based on image complexity and clinical relevance, which significantly reduces inference costs without sacrificing diagnostic fidelity.
    \item  We conduct extensive experiments on diverse pathological tasks, demonstrating consistent improvements in both accuracy and efficiency.
\end{itemize}

\section{Related Works}

\subsection{Multimodal Computational Pathology}
The field of computational pathology has witnessed a surge in foundation models (FMs)~\cite{uni,conch,gpfm,mstar,huang2023visual,xiang2025vision} aimed at enhancing diagnostic precision and prognostic evaluation.
Building upon these FMs, the integration of large language models (LLMs) has catalyzed the emergence of MLLMs \cite{liu2023visual,liu2024improved,bai2025qwen2,pathasst}, which demonstrate significant advancements in addressing complex, open-world visual tasks. 
These models hold strong potential to serve as flexible assistants across various aspects of medical practice, including clinical decision support, medical education, and biomedical research.
Quilt-LLaVa~\cite{quiltllava} builds upon the Quilt-Instruct dataset, which contains over 107K pathology instruction–answer pairs grounded in diagnostically relevant WSI patches, and enables cross-patch diagnostic reasoning on whole-slide images (WSIs).
PathChat~\cite{pathchat} is a vision-language AI assistant tailored for pathology, trained on over 450K visual-language instruction pairs. 
By integrating a pathology FM with a LLM, PathChat achieves strong performance on both multiple-choice and open-ended diagnostic tasks, outperforming general-purpose models like GPT-4V.
However, all these methods exhibit limited reasoning capabilities and are computationally inefficient, restricting their practical deployment.

\subsection{Reinforcement Learning for Reasoning}
RL has shown great potential in enhancing the reasoning capabilities of LLMs through reward-driven optimization since the release of reasoning models like OpenAI’s o1~\cite{openaio1}.
A significant advancement in this area is Deepseek-R1~\cite{deepseekr1}, which demonstrates that strong reasoning capabilities can be achieved through reinforcement learning alone, bypassing the need for the SFT stage.
Visual-RFT~\cite{visualrft} extends RL to the natural images by leveraging verifiable visual rewards to optimize MLLMs for tasks such as fine-grained classification. 
It introduces a reward-driven framework that improves reasoning and generalization under limited supervision, offering an alternative to traditional SFT. 
Med-R1~\cite{med-r1} applies reinforcement learning to enhance MLLMs for medical imaging tasks, addressing challenges posed by limited annotations and the need for clinically coherent reasoning. 
To the best of our knowledge, we are the first to jointly tackle the problem of  pathological understanding and token allocation using reinforcement learning, marking a significant advancement toward more efficient and clinically viable computational pathology systems.

%%%%%%%%%%%%%%%%%%%%%%%%%%%%%%%%%%%%%%%%%%%%%%%%%%%%%%%%%%%%%%%%%%%%%%%%%%%%%%%%%%%
\section{Methodology}
\subsection{Preliminaries}
\subsubsection{Group Relative Policy Optimization}
Group relative policy optimization (GRPO) is a variant RL algorithm of proximal policy optimization (PPO) \citep{schulman2017proximal}, which foregoes the critic model and instead estimates the baseline from group scores, significantly reducing training resources. Specifically, for each question $q$, GRPO samples a group of outputs $\{o_1, o_2, \cdots, o_G\}$ from the old policy $\pi_{\theta_{old}}$ and then optimizes the policy model $\pi_{\theta}$ by maximizing the following objective:
\begin{equation}
\begin{split}
    \mathcal{J}&_{GRPO}(\theta) = \mathbb{E}{[q \sim P(Q), \{o_i\}_{i=1}^G \sim \pi_{\theta_{old}}(O|q)]}  \\
    & \frac{1}{G}\sum_{i=1}^G \left( \min \left( \frac{\pi_\theta(o_i |q)}{\pi_{\theta_{old}}(o_i |q)} A_i, \text{clip} \left( \frac{\pi_\theta(o_i |q)}{\pi_{\theta_{old}}(o_i |q)}, 1 - \epsilon, 1 + \epsilon \right)  A_i \right) - \beta \mathbb{D}_{KL}\left(\pi_{\theta} || \pi_{ref}\right)\right) ,
\end{split}
\label{eq:GRPO-obj}
\end{equation}
\begin{equation}
    \mathbb{D}_{KL}\left(\pi_{\theta} || \pi_{ref}\right) = \frac{\pi_{ref}(o_i|q)}{\pi_{\theta}(o_i|q)}- \log\frac{\pi_{ref}(o_i|q)}{\pi_{\theta}(o_i|q)} - 1,
\end{equation}
where $\epsilon$ and $\beta$ are hyper-parameters, and $A_i$ is the advantage, computed using a group of rewards $\{r_1, r_2, \ldots, r_G\}$ corresponding to the outputs within each group:
\begin{equation}
    A_i = \frac{r_i - {\operatorname{mean}(\{r_1, r_2, \cdots, r_G\})}}{{\operatorname{std}(\{r_1, r_2, \cdots, r_G\})}}.
\end{equation}

\subsubsection{Visual Tokenization}
Typically, MLLM employs a patch-wise tokenization process to convert input images into a sequence of visual tokens. Given an input image $\mathbf{I} \in \mathbb{R}^{H \times W \times C}$ with height $H$, width $W$, and $C$ channels, it is first divided into $N$ non-overlapping patches of size $P \times P$ where $N = H/P \times W/P$. Each patch is flattened into a 1D vector $\mathbf{x}_p \in \mathbb{R}^{P^2C}$ and then mapped to $D$-dimensional embedding space via a learnable projection matrix $\mathbf{E} \in \mathbb{R}^{(P^2C) \times D}$, \emph{i.e.}, $\mathbf{z}_p = \mathbf{x}_p\mathbf{E} + \mathbf{e}_p^{pos}$, where $\mathbf{e}_p^{pos} \in \mathbb{R}^D$ represents the positional embedding for patch $p$.
The resulting token sequence $\mathbf{Z} \in \mathbb{R}^{N \times D}$ serves as input to the transformer encoder, where $D$ matches the model's hidden dimension.
\begin{figure}
    \centering
    \includegraphics[width=\linewidth]{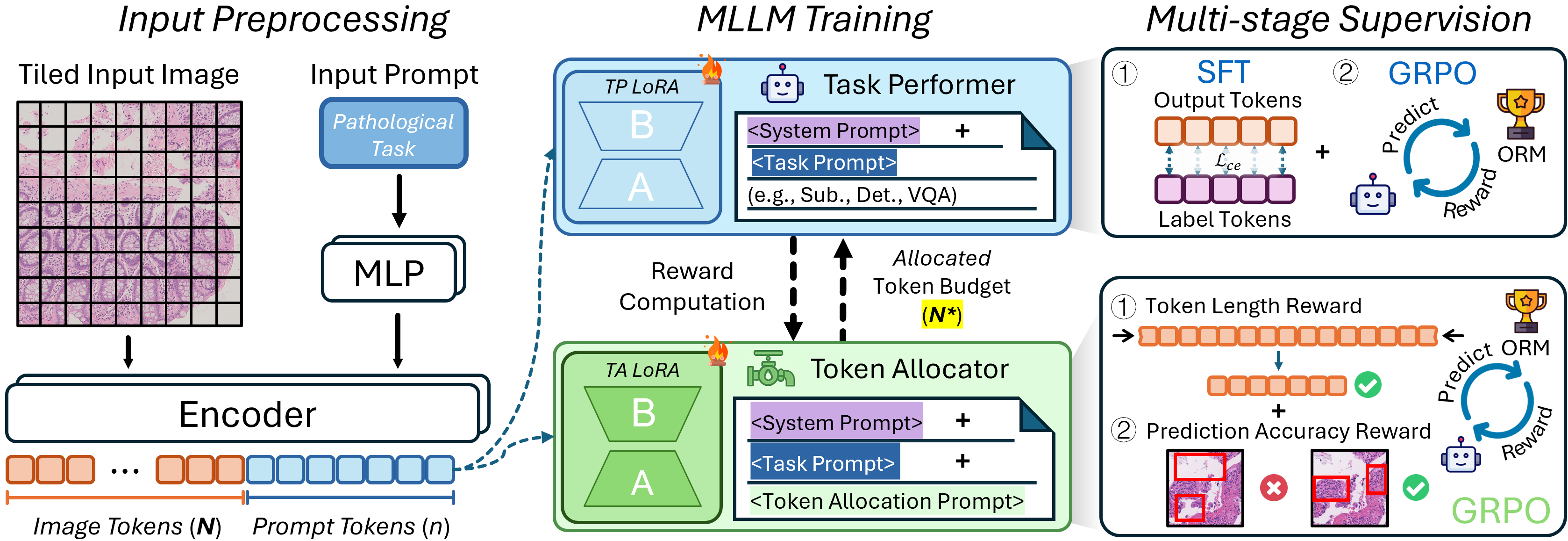}
    \caption{Overall framework of our method. High-resolution pathological images are first preprocessed into tiled patches and, together with task prompts, encoded into tokens. A task performer processes these tokens under SFT and GRPO supervision, while a token allocator dynamically adjusts the token budget via reinforcement learning.}
    \label{fig:scheme}
\end{figure}

\subsection{Bilateral Pathological Reasoning}
Fig. \ref{fig:scheme} presents the overall framework of our method for effective pathological reasoning. The input image and the accompanying text prompt are first encoded into tokens with a limited budget. The task performer then processes these tokens under a multi-stage supervision mechanism, including SFT and RL with task-specific rewards. A token allocator optimally distributes a token budget across task prompts and images, ensuring efficient resource utilization.  This integrated approach enables flexible handling of diverse pathology vision-language tasks (e.g., VQA, cancer subtyping, and lesion detection) while balancing computational efficiency and task performance through dynamic token allocation and hybrid optimization strategies.

\subsubsection{Pathological Reinforcement Training with Limited Token Budget}
The analysis of high-resolution pathological images, especially using MLLM, presents a significant challenge due to their exceptionally large size. For example, even dividing a $2048\times2048$ region image from a histopathological image into $28 \times 28$ non-overlapping patches would generate over $5,000$ tokens. Moreover, training requires storing gradients, optimizer states, and intermediate activations for all tokens in the sequence for backpropagation, which consumes significantly more memory and time than inference. To this end, inspired by the dynamic resolution of Qwen2.5-VL models \cite{bai2025qwen2}, we apply a resize operation $\mathscr{T}^{M,P}: \mathbb{R}^{H \times W \times 3} \rightarrow \mathbb{R}^{H' \times W' \times 3}$ to process images of varying sizes and limit the number of training tokens per image to $M$:
\begin{equation}
H' \equiv 0 \pmod{P}, \quad
W' \equiv 0 \pmod{P}, \quad 
H'W' \leq M P^2,
\label{equ:transform}
\end{equation}
where $P$ is the patch size. 

The scaling factor $\gamma$ is computed as:
\begin{equation}
\gamma = \begin{cases}
\sqrt{\dfrac{N}{M P^2}} & \text{if } \left\lceil\dfrac{N}{P^2}\right\rceil > M, \\
1 & \text{otherwise}.
\end{cases}
\end{equation}

The final dimensions are determined by \(P\)-aligned rounding: 
\begin{equation}
H' = P \cdot \left\lfloor\dfrac{H}{\gamma P}\right\rfloor, \quad 
W' = P \cdot \left\lfloor\dfrac{W}{\gamma P}\right\rfloor.
\end{equation}
which ensures that $H'$ and $W'$ are exact multiples of $P$ while maintaining the aspect ratio within $\pm P$ pixels. Given the original image $\mathbf{I}_{\text{rgb}}$, the output image $\mathbf{I} = \mathscr{T}^{M,P}(\mathbf{I}_{\text{rgb}})$ satisfies all architectural requirements for visual tokenization.

Current multimodal approaches \cite{sun2024cpath,xiang2025vision,pathchat,pathasst} for pathological image analysis remain constrained to simple tasks and demonstrate limited reasoning capacity, failing to address the nuanced demands of complex diagnostic workflows. To bridge this gap, our methodology adopts a two-phase paradigm to train task performer with low-rank adaptation (LoRA) \cite{hu2022lora}. First, we perform SFT to establish robust feature representations and task-specific baselines. This initial phase ensures the model acquires fundamental diagnostic competencies across diverse pathological patterns before advancing to more sophisticated reasoning. Building upon this foundation, we implement reinforcement learning through GRPO, inspired by the success of DeepSeek-R1 \cite{deepseekr1} in enhancing reasoning capabilities. 
We propose specialized reward functions tailored to distinct pathological tasks, including VQA, cancer subtyping, and lesion detection. The pathological reasoning prompts for different tasks, including system prompt $\mathsf{p^{sys}}$, VQA prompt $\mathsf{p^{vqa}}$, subtyping prompt $\mathsf{p^{sub}}$, detection prompt $\mathsf{p^{det}}$, and token allocation prompt $\mathsf{p^{ta}}$, are shown in the supplementary material \ref{app:templates}. 

For each task, the model generates a text response $\mathsf{r}$ which is then parsed to extract task-relevant outputs. All rewards combine task performance ($R_{\text{task}}$) and format compliance ($R_{\text{format}}$):
$R = R_{\text{task}} + \lambda R_{\text{format}}$, where $\lambda$ controls the format penalty weight (set to 1).
The format reward $R_{\text{format}}$ enforces the model to put its thinking process and answer between ‘<think>’ ‘</think>’ and ‘<answer>’ ‘</answer>’ tags, respectively.

\textbf{VQA Reward in VQA Tasks.}
Given an input image $\mathbf{I}$, system prompt $\mathsf{p^{sys}}$, and VQA prompt $\mathsf{p^{vqa}}$, the Task Perfomer $\operatorname{TP}$ generates response $\mathsf{r} = \operatorname{TP}(\mathbf{I}, \mathsf{p^{sys}}, \mathsf{p^{vqa}})$ with the answer $\hat{\mathsf{a}}$ parsed using regular expressions.
The VQA reward is computed as:
$R_{\text{vqa}}(\mathsf{r}, \hat{\mathsf{a}},\mathsf{a}) = R_{\text{ans}}(\hat{\mathsf{a}}, \mathsf{a}) + \lambda R_{\text{format}}(\mathsf{r})$.
The answer reward $R_{\text{ans}}(\hat{\mathsf{a}}, \mathsf{a})$ differs for close-ended and open-ended questions:
\begin{equation}
R_{\text{ans}}(\hat{\mathsf{a}}, \mathsf{a}) = 
\begin{cases}
\mathbb{I}(\hat{\mathsf{a}} = \mathsf{a}) & \text{(close-ended)} \\
\text{BLEU}(\hat{\mathsf{a}}, \mathsf{a}) & \text{(open-ended)}
\end{cases}
\end{equation}
where $\mathsf{a}$, $\mathbb{I}(\cdot)$ and $\text{BLEU}(\cdot)$ are the ground truth answer, the indicator function, and the BLEU score, respectively.

\textbf{SUB Reward in Subtyping Tasks.}
In cancer subtyping tasks, we define a reward function $R_{\text{sub}}(\mathsf{r}, \hat{\mathsf{y}},\mathsf{y}) = R_{\text{acc}}(\hat{\mathsf{y}},\mathsf{y}) + \lambda R_{\text{format}}(\mathsf{r})$, where $\mathsf{r} = \operatorname{TP}(\mathbf{I}, \mathsf{p^{sys}}, \mathsf{p^{sub}})$ with the disease category $\hat{\mathsf{y}}$. Specifically, the accuracy reward $R_{\text{acc}}$ is given by the indicator function:
\begin{equation}
R_{\text{acc}}(\hat{\mathsf{y}}, \mathsf{y}) = \mathbb{I}(\hat{\mathsf{y}} = \mathsf{y}) = 
\begin{cases} 
1 & \text{if } \hat{\mathsf{y}} = \mathsf{y} \\
0 & \text{otherwise}
\end{cases}
\end{equation}
where $\mathsf{y} \in \mathcal{Y}$ is the ground truth label.

\textbf{DET Reward in Detection Tasks.}
For lesion detection, we construct a composite reward $R_{\text{det}}$ that jointly considers both detection performance and output standardization, \emph{i.e.}, $R_{\text{det}}(\mathsf{r}, \mathcal{P}, \mathcal{G}) =  R_{\text{AP}}(\mathcal{P}, \mathcal{G}) + \lambda R_{\text{format}}(\mathsf{r})$. The Average Precision (AP) reward $R_{\text{AP}}$ is computed as:
\begin{equation}
R_{\text{AP}}(\mathcal{P}, \mathcal{G}) = \text{AP}_{50}(\mathcal{P}, \mathcal{G}), 
\end{equation}
where $\mathcal{P} = \{\mathbf{p}_i\}_{i=1}^I$ is the set of predicted bounding boxes parsed from the response $\mathsf{r} = \operatorname{TP}(\mathbf{I}, \mathsf{p^{sys}}, \mathsf{p^{det}})$ and $\mathcal{G} = \{\mathbf{g}_j\}_{j=1}^J$ is the set of ground truth boxes. $\text{AP}_{50}$ is Average Precision at IoU threshold 0.5. 

\subsubsection{Task-Dependent Token Allocation}

While the task performer demonstrates promising pathological reasoning capabilities, we identify a critical limitation in the static token allocation strategy. Empirical analysis reveals that static token budgets result in suboptimal efficiency-accuracy trade-offs across diverse diagnostic scenarios (see examples in  \ref{app:discussion}). Specifically, static allocation often leads to over-provisioning in routine cases, such as those involving large tumor regions or clear inflammatory patterns, where accurate interpretation can be achieved with significantly fewer tokens, thereby wasting computational resources. Conversely, in complex diagnostic cases that require capturing fine-grained cellular interactions, rigid token limits can force premature feature compression, ultimately constraining model accuracy.
    
This motivates our dynamic token allocation method, which enables adaptive resource distribution based on image complexity and task demands. To achieve this automatically without manual heuristics, we formulate token allocation as another reinforcement reasoning problem where a token allocator $\operatorname{TA}$ is trained with LoRA to optimize the trade-off between computational efficiency and diagnostic accuracy.
Specifically, we define a composite reward function $R_{\text{TA}}(\mathsf{r},N_p,N_0) =  R_{\text{token}}(N_p, N_0) + \lambda R_{\text{format}}(\mathsf{r})$.
The token allocation reward $R_{\text{token}}$ is defined as:
\begin{equation}
R_{\text{token}}(N_p, N_0) = 
\begin{cases} 
R_{\text{task}} & \text{if }  N_p \leq N_0 \\
\alpha*R_{\text{task}} & \text{if }  N_p > N_0 \\
\end{cases}
\end{equation}
where $\mathsf{r} = \operatorname{TA}(\mathbf{I}, \mathsf{p^{sys}}, \mathsf{p^{ta}})$. $N_p$ is the predicted token count and $N_0$ is the original token count (less than the token budget $M$ as mentioned before). $\alpha$ is a coefficient less than 1 (set to 0.5). $R_{\text{task}}$ represents $R_{\text{ans}}$,$R_{\text{acc}}$, and $R_{\text{AP}}$ for VQA, subtyping, and detection tasks, respectively.
This formulation encourages token reduction when performance gains are limited, while permitting token increases when matched by sufficient performance improvement. Such dynamic allocation allows the model to automatically balance token usage, conserving resources for easier cases and allocating more tokens for diagnostically challenging ones. The task performer, with allocated token number \(N_p\) from the token allocator, is used to compute \(R_{\text{task}}\).

\section{Experiments}

\subsection{Datasets and Evaluation Protocol}

\textbf{Datasets.} 
We evaluate our method on six public datasets: PathMMU \cite{sun2024pathmmu}, PathVQA \cite{he2020pathvqa}, UniToPatho \cite{barbano2021unitopatho}, ESCA \cite{tolkach2023artificial}, CRAG \cite{graham2019mild}, and DigestPath2019 Tissue \cite{da2022digestpath}. 
Among these datasets, PathMMU and PathVQA are utilized for VQA tasks, UniToPatho and ESCA are for subtyping, while CRAG and DigestPath2019 Tissue are for lesion detection. Details of the datasets are in the supplementary material \ref{app:datasets}.

\textbf{Evaluation Metrics.}
Following previous works \cite{liu2020deep,papineni2002bleu}, we employ task-specific evaluation metrics to comprehensively assess model performance across different tasks. For VQA tasks, we distinguish between close-ended and open-ended questions: for close-ended questions, we report standard accuracy (\emph{ACC}), while for open-ended questions, we employ \emph{BLEU} score to evaluate semantic alignment between predicted and ground-truth answers. For subtyping tasks, we utilize balanced accuracy (\emph{Bal-ACC}) to address potential class imbalance and \emph{W-F1} score to balance precision and recall across all categories. Detection performance is evaluated using mean average precision (\emph{mAP}) calculated over intersection over union (\textit{IoU}) thresholds ranging from 0.1 to 0.5 with an interval of 0.2, providing a robust measure of localization accuracy. In addition, tokens per image (TPI) is utilized to measure the computational efficiency.

\begin{table}[t]
	\centering{
	\caption{Overall results of models on the PathMMU \textbf{validation} and \textbf{test set}. The best-performing methods are \textbf{bolded} and second-best ones are \underline{underlined}. Results of Human and MLLMs in the middle are from \cite{sun2024pathmmu}. - denotes not applicable.}\label{tab:pathmmu}
    \resizebox{\linewidth}{!}{
	\begin{tabular}{@{}lccccccccccc@{}}
			\toprule
			Subset $\rightarrow$ &\textbf{Test} &\textbf{Val} & \multicolumn{2}{c}{\textbf{PubMed}}  & \multicolumn{2}{c}{\textbf{EduContent}} & \multicolumn{2}{c}{\textbf{Atlas}} &\multicolumn{2}{c}{\textbf{PathCLS}} &\\ 
			 Method $\downarrow$ &\textbf{Overall} &\textbf{Overall} & Test  & Val     & Test  & Val  & Test  & Val  & Test  & Val &TPI\\\midrule
			% Method $\downarrow$&(921) & (555)  & (281)  & (233)  & (255)  & (146)  & (208)  & (80) & (177)  & (96) 
			% \color{gray} Random Choice & \color{gray} 21.2 & \color{gray}    & \color{gray} 22.1 & \color{gray}-  & \color{gray} 25.5 & \color{gray} -& \color{gray} 19.7 & \color{gray} -  & \color{gray} 15.3 & \color{gray} - &\\ 
			% \color{gray} Frequent Choice & \color{gray} 27.6  & \color{gray}   & \color{gray} 28.8 & \color{gray}   & \color{gray} 29.8 & \color{gray}  & \color{gray} 28.4 & \color{gray}  & \color{gray} 22.0 & \color{gray} &  \\	
 Human Expert & \underline{72.0} & -   &  72.9 & - &  69.0 &  - & 68.3 &  - & \textbf{78.9} &  - & -\\
% \midrule
% Qwen-VL-7B               &31.6  &   & 34.9 & -  & 34.1 & - & 35.6 & - & 18.6 & -  &881.5\\			
% BLIP-2 FLAN-T5-XXL       &32.2  & & 37.0 & -  & 30.2 & - & 39.4 & - & 19.8 & -  &881.5\\
InstructBLIP-FLAN-T5-XXL &34.4  &-   & 39.1 & - & 34.5 & - & 38.5 & - & 22.6 & -  &881.5\\
LLaVA-1.5-13B            & 38.4 & - & 44.5 & -& 34.1 & - & 47.1 & - & 24.9 & -  &881.5\\
Qwen-VL-MAX        &48.0  & -  & 53.0 & -  & 52.2 & -  & 51.4 & - & 30.5 & - &881.5\\
Gemini Pro Vision  & 42.9 & -  & 43.8 & - & 43.5 & - & 49.5 &- & 32.8 & -  &881.5 \\
GPT-4V             & 52.7 &  - & 59.4 & -  & 60.4 & - & 48.1 & -  & 36.2 & - &881.5\\   
            \midrule
Qwen2.5-VL-7B            & 48.0 &39.8    &55.5  &42.1   & 52.9 & 41.8  & 46.2 & 46.3& 31.6 & 26.0  &881.5 \\	
SFT            &67.1  & 60.0   &66.5  &54.1   & 69.4 & 61.6  & \underline{71.1} & \textbf{75}& 60.5 & 60.4  &\underline{239.9} \\	
\rowcolor{gray!15} +Performer (Ours)&\textbf{72.2}  & \underline{65.0}   &\textbf{74.7}  &\underline{62.2}  & \underline{73.3} & \underline{61.6}  & \textbf{72.1} & \underline{73.8}& 67.2 & \textbf{70.8}  &\underline{239.9} \\	
\rowcolor{gray!15} +Performer+Allocator (Ours)&\textbf{72.2}  & \textbf{65.9}   &\underline{73.0}  &\textbf{64.8}  & \textbf{74.9} & \textbf{63.7}  & 69.2 & \underline{73.8}& \underline{70.6} & \underline{65.6}  &\textbf{112.7} \\	
			\bottomrule
		\end{tabular}
}
}
\end{table}

\begin{table}
\centering
\caption{Overall results of models on the PathVQA dataset.}\label{tab:pathavqa}%
% \resizebox{\linewidth}{!}{
\footnotesize
\begin{tabular}{lcccc}
\toprule
  Method & ACC (Close) & BLEU (Open) &TPI &Reasoning\\
	\midrule
Qwen2.5-VL-7B & 49.0 & 0.001   & 493.9 & \ding{55} \\
SFT &85.4  & 0.181     & \underline{250.0} & \ding{55} \\
\rowcolor{gray!15}+Performer (Ours)&\underline{86.9}  & \underline{0.197}   & \underline{250.0} & \ding{52} \\
\rowcolor{gray!15}+Performer+Allocator (Ours)&\textbf{89.2}  & \textbf{0.205}    &\textbf{195.4} & \ding{52} \\
 \bottomrule
\end{tabular}
% }
\end{table}

\subsection{Implementation Details}
Our method uses Qwen2.5VL-7B \cite{bai2025qwen2} as the base model, which is currently one of the best-performing open-source large multimodal models. We set the token budget $M$ and patch size $P$ to 256 and 28, respectively. The intrinsic rank and global scaling factor in LoRA are set to 16 and 64. AdamW \cite{kingma2014adam} is used as the optimizer with a weight decay of 0.1. The initial learning rate is set to 1e-4 with a cosine learning rate schedule. KL regularization term coefficient $\beta$, clip coefficient $\epsilon$, and the number of sampled outputs $G$ in a group for GRPO are set to 0.001, 0.2, and 8, respectively. The model was implemented with PyTorch \cite{paszke2019pytorch} and trained on a 8$\times$64GB MetaX MXC500 GPU node. More details of the implementation for baseline methods can be found in the supplementary material \ref{app:implementation}.

\subsection{Experimental Results and Analysis}
\textbf{Visual Question Answering Tasks.}
As shown in Tab. \ref{tab:pathmmu} and \ref{tab:pathavqa}, the experimental results on the PathMMU and PathVQA demonstrate a clear hierarchy of performance. 
The MLLMs show varying degrees of competency, with GPT-4V emerging as the strongest off-the-shelf model, followed closely by Qwen-VL-MAX  and Gemini Pro Vision. 
The base Qwen2.5-VL-7B shows competitive performance without specialized training. SFT yields dramatic improvements.
Our task performer significantly pushes performance, while token allocator further improves the performance with lower image tokens (12.8\% of base model and 47.0\% of task performer).
The consistent improvements across all subsets validate our approach's effectiveness and efficiency in bridging the gap between general multimodal models and specialized pathological image analysis.
Notably, our best-performing model can outperform human expert on the PubMed, EduContent, and Atlas subsets of PathMMU.

\textbf{Subtyping Tasks.}
Tab. \ref{tab:unitopatho} shows the results on the UniToPatho and ESCA datasets, respectively. 
Among traditional pathology-specialized models, GPFM achieves the highest performance with Balanced ACC and  Weighted F1, closely followed by Prov-Gigapath and PLIP. Notably, all these models leverage pathology-specific pre-training but lack explicit reasoning capabilities, indicating that domain-adaptive pre-training alone, while beneficial, may not suffice for complex diagnostic tasks requiring higher-order reasoning.
The base Qwen2.5-VL-7B model without pathological pre-training or reasoning performs poorly, highlighting the challenge of transferring general-domain vision-language models to pathological tasks. The results from our proposed methods demonstrate progressive improvements through successive training stages. With SFT, performance improves, suggesting that task-specific adaptation can partially compensate for the lack of pathological pre-training. Further enhancements are observed with our task performer and token allocator, which introduce reasoning capabilities and computational efficiency. This underscores the effectiveness of our method in bridging the reasoning gap without relying on expensive pathological supervision.

\begin{table}[t]
\centering
\caption{Subtyping performance of different models on the UniToPatho and ESCA datasets. * denotes methods with large-scale pathology pre-training. - denotes not applicable.
}\label{tab:unitopatho}%
% \resizebox{\linewidth}{!}{
\footnotesize
\begin{tabular}{lccccccc}
\toprule
Dataset $\rightarrow$ & \multicolumn{3}{c}{\textbf{UniToPatho}} & \multicolumn{3}{c}{\textbf{ESCA}}\\
  Method $\downarrow$ & Bal-ACC & W-F1 & TPI & Bal-ACC & W-F1 & TPI &Reasoning \\
	\midrule
 ResNet50  &  39.7 &  38.4 & -  & 60.1 & 55.3 &-& \ding{55} \\
 Phikon*  & 37.9 &  37.5 & - & 66.8 & 64.2 &- & \ding{55}\\
 Ctranspath*  &  31.0 &  30.2 & -& 64.2 & 66.0 &-  & \ding{55}\\
 % UNI  & \underline{46.2 }& \underline{45.5 } & -  & \ding{52}\\
 % CONCH &  \pmb{52.2 }& \pmb{52.7 } & - & \ding{52} \\
 PLIP* & 43.7 & 41.8  & - & 60.1 & 55.2 &- & \ding{55} \\
 CHIEF*& 39.4 & 38.6  & - & 60.9 & 62.8 &- & \ding{55}\\
Prov-Gigapath*& \underline{44.2} & 43.7 & - & \underline{72.5} & 73.8 & - & \ding{55}\\
GPFM*& \textbf{44.4} &  43.3  & -& \textbf{73.2} & 73.4 & -&\ding{55}\\
	\midrule
Qwen2.5-VL-7B & 14.7 & 8.6 &  4225.0 & 15.9 & 28.2 &  \underline{81.0} & \ding{55}\\
SFT & 36.9 &  43.2  & \underline{256.0} & 65.7 &  81.7  & \underline{81.0} & \ding{55}\\
\rowcolor{gray!15}+Performer (Ours)& 42.5 &  \underline{46.8}  & \underline{256.0} & 68.2 &  \underline{87.3}  &  \underline{81.0} & \ding{52}\\
\rowcolor{gray!15}+Performer+Allocator (Ours)& 42.3 & \textbf{47.5}   & \textbf{251.5} & 69.6 & \textbf{88.5}   & \textbf{76.4} & \ding{52}\\
 \bottomrule
\end{tabular}
% }
\end{table}

\textbf{Detection Tasks.}
We present the results on the DigestPath2019 Tissue and CRAG datasets in Tab. \ref{tab:dg19tissue} and Tab. \ref{tab:crag}, respectively. The baseline Qwen2.5-VL-7B model shows limited detection capability, achieving only 10.6\% and 11.4\% average mAP across IoU thresholds from 0.1 to 0.5, with particularly poor performance at the stringent $IoU=0.5$ criterion.
SFT brings dramatic improvements, which highlight the importance of task-specific adaptation.
Our proposed method further elevates performance to a new state-of-the-art level. The overall average mAP of represents a more than 37.2\% and 26.1\% boost over the SFT baseline and an 7$\times$ improvement over the original model. 
The results strongly support the effectiveness of our approach in bridging the gap between general vision-language models and specialized pathological image analysis tasks.

\begin{table}
\centering{
\caption{Detection performance of different models on the DigestPath2019 Tissue dataset. }\label{tab:dg19tissue}%
    % \resizebox{\linewidth}{!}{
\footnotesize
\begin{tabular}{l c c c c c c }
\toprule
Method &mAP@0.1 &mAP@0.3 &mAP@0.5 &AVG &TPI &Reasoning\\
\midrule
Qwen2.5-VL-7B &22.6 &7.1 &2.2 &10.6 &1369.0&\ding{55} \\
SFT &59.3 &54.1 &46.9 &53.4 &\underline{256.0} &\ding{55}\\
\rowcolor{gray!15}+Performer (Ours) &\underline{77.1} &\underline{71.0} &\underline{61.6} &\underline{69.9} &\underline{256.0} &\ding{52}\\
\rowcolor{gray!15}+Performer+Allocator (Ours)&\textbf{82.1}&\textbf{74.7}&\textbf{63.0}&\textbf{73.3}&\textbf{244.0}&\ding{52}\\
\bottomrule
\end{tabular}
}
% }
\end{table}

\begin{table}
\centering{
\caption{Detection performance of different models on the CRAG dataset. }\label{tab:crag}%
    % \resizebox{\linewidth}{!}{
\footnotesize
\begin{tabular}{l c c c c c c}
\toprule
Method &mAP@0.1 &mAP@0.3 &mAP@0.5 &AVG &TPI &Reasoning\\
\midrule
Qwen2.5-VL-7B &24.8 &7.6 &1.9 &11.4 &1369.0 &\ding{55}\\
SFT &62.3 &59.2 &54.2 &58.6 &\underline{256.0} &\ding{55}\\
\rowcolor{gray!15}+Performer (Ours) &\underline{77.0} &\underline{73.3} &\textbf{66.0} &\underline{72.1} &\underline{256.0} &\ding{52}\\
\rowcolor{gray!15}+Performer+Allocator (Ours)&\textbf{81.7}&\textbf{75.6}&\underline{64.5}&\textbf{73.9}&\textbf{229.8} &\ding{52}\\
\bottomrule
\end{tabular}
}
% }
\end{table}

\begin{table}
	\centering{
	\caption{Comparison of token allocator with fixed tokenization strategy using different token budget.}\label{tab:ablation}
    \resizebox{0.95\linewidth}{!}{
	\begin{tabular}{@{}lccccccccccc@{}}
			\toprule
			% Subset $\rightarrow$ & \textbf{\begin{tabular}[c]{@{}c@{}}Test \\  Overall\end{tabular}}& \textbf{\begin{tabular}[c]{@{}c@{}}Val \\  Overall\end{tabular}} & \multicolumn{2}{c}{\textbf{PubMed}}  & \multicolumn{2}{c}{\textbf{EduContent}} & \multicolumn{2}{c}{\textbf{Atlas}} &\multicolumn{2}{c}{\textbf{PathCLS}} &\\ 
            Subset $\rightarrow$ &\textbf{Test} &\textbf{Val} & \multicolumn{2}{c}{\textbf{PubMed}}  & \multicolumn{2}{c}{\textbf{EduContent}} & \multicolumn{2}{c}{\textbf{Atlas}} &\multicolumn{2}{c}{\textbf{PathCLS}} &\\ 
			 Method $\downarrow$ &\textbf{Overall} &\textbf{Overall} & Test  & Val     & Test  & Val  & Test  & Val  & Test  & Val &TPI\\\midrule
\rowcolor{gray!15} Performer (Ours, $\hat{M}=256$)&\textbf{72.2}  & \underline{65.0}   &\textbf{74.7}  &\underline{62.2}  & \underline{73.3} & 61.6 & \textbf{72.1} & \textbf{73.8}& \underline{67.2} & 70.8  &239.9 \\	
$\hat{M}=128$&68.5&58.4&71.2&57.9&69.8&54.8&69.2&\underline{67.5}&62.1&58.3&\underline{123.6} \\	
$\hat{M}=512$&68.5&62.5&70.8&62.2&68.6&54.8&69.2&63.7&64.4&\textbf{75}&420.7 \\	
$\hat{M}=1024$&69.2&64.7&69.0&61.4&72.5&\underline{63.0}&\underline{70.2}&\underline{67.5}&63.8&\underline{72.9}&613.3 \\	
\rowcolor{gray!15}+Allocator (Ours)&\textbf{72.2}  & \textbf{65.9}   &\underline{73.0}  &\textbf{64.8}  & \textbf{74.9} & \textbf{63.7}  & 69.2 & \textbf{73.8}& \underline{70.6} & 65.6  &\textbf{112.7} \\	
			\bottomrule
		\end{tabular}
}
}
% \vspace{-0.5cm}
\end{table}

\begin{figure}
    \centering
    \includegraphics[width=\linewidth]{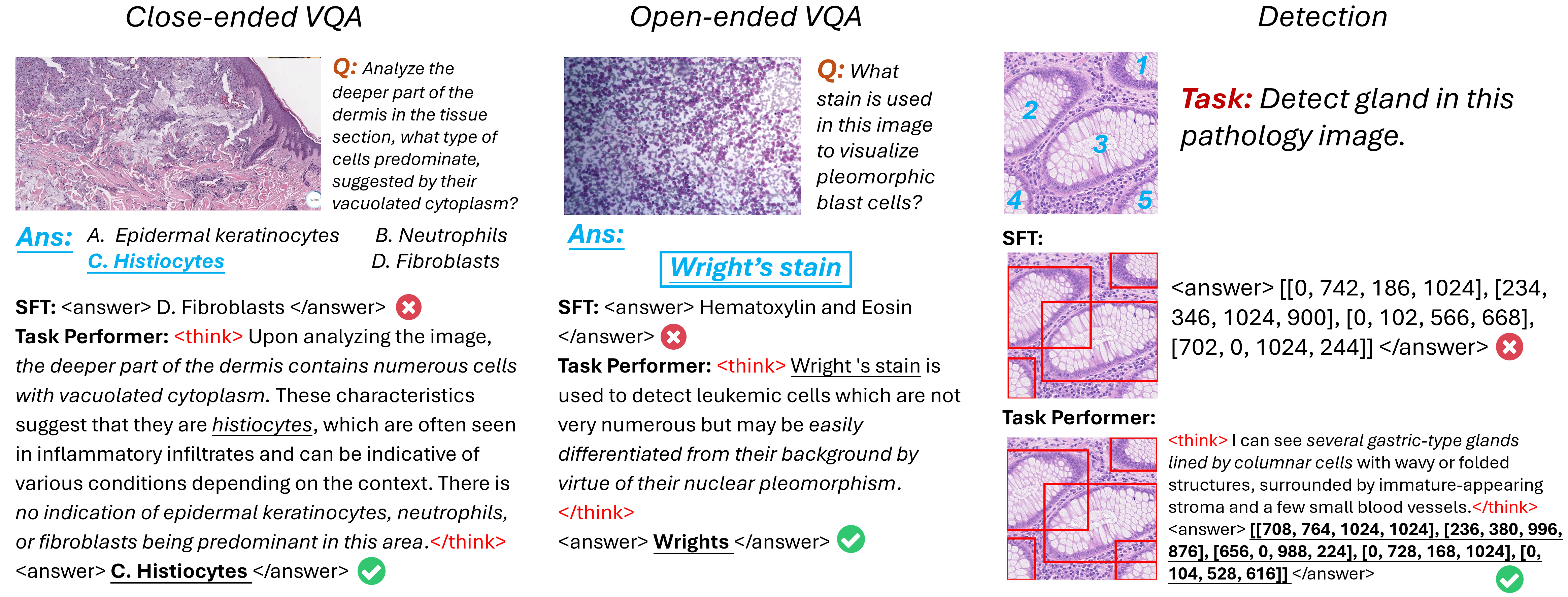}
    \caption{Qualitative results of pathological reasoning. }
    \label{fig:qualitative}
\end{figure}

\textbf{Representative qualitative results of pathological reasoning.}
As shown in Fig. \ref{fig:qualitative}, our task performer yields more accurate and interpretable outcomes compared to SFT across multiple tasks, including VQA, subtyping, and detection. Specifically, the task performer exhibits structured reasoning, providing intermediate pathology rationales before arriving at a final answer. For example, in the first task, the task performer first identifies cellular morphology (vacuolated cytoplasm) before concluding the cell type (histiocytes). SFT models, in contrast, tend to produce direct answers without explicit justification, making their predictions less transparent and harder to validate. More results can be found in the supplementary material \ref{app:qualitative}.

\textbf{Effectiveness of the proposed token allocator.}
To verify the effectiveness of the token allocator, we conduct experiments to directly limit the max token to $\hat{M}=128/512/1024$ during testing. Images with original token number greater than $\hat{M}$ will be downsampled, and those with original token number less than $\hat{M}$ will basically remain unchanged, as shown in Eq. \ref{equ:transform}. We present the performance of various methods in Tab. \ref{tab:ablation}. The metrics reported are accuracy scores and the tokens per image (TPI).
Our method with token allocator stands out by achieving high accuracy with the lowest token usage (TPI=112.7), demonstrating superior efficiency. This suggests that adaptive token allocation can optimize resource utilization while maintaining performance.
In contrast, $\hat{M}=512$ and $\hat{M}=1024$ use more tokens but do not consistently outperform the baseline or token-allocation variant, highlighting the importance of strategic token management. $\hat{M}=128$ performs poorly, suggesting that overly restrictive token limits may harm performance on complex tasks.

\textbf{Limitations.}
Despite the promising results, our study has several limitations that warrant further consideration. First, all experiments used 7B-parameter models due to computational constraints, which may cap performance compared to larger models. Second, resource limitations restricted our analysis to large ROIs rather than WSIs. 
Finally, further validation on more diverse and larger-scale pathological datasets is needed. Future work could address these issues by employing more efficient architectures \cite{yang2024mambamil, nasiri2024vim4path} and extending to whole-slide analysis to improve robustness and clinical relevance \cite{ferber2024context}.

\section{Conclusion}
In this work, we present a novel bilateral reinforcement learning framework to address the challenges of limited reasoning capability and computational inefficiency in pathological image analysis. By integrating two reinforcement branches, one enhancing reasoning through indirect supervision and another optimizing token allocation, our approach significantly improves diagnostic performance while reducing computational overhead. Extensive experiments on visual question answering, cancer subtyping, and lesion detection demonstrate that the proposed framework achieves a 41.7 overall performance gain while cutting average token costs by 70.3\% compared to the original base model. These results highlight the framework’s scalability and its potential for real-world clinical deployment, advancing both multimodal reasoning and practical AI-assisted diagnostics in pathology.

\bibliographystyle{plainnat}
\bibliography{bibliography.bib}

%%%%%%%%%%%%%%%%%%%%%%%%%%%%%%%%%%%%%%%%%%%%%%%%%%%%%%%%%%%%

\newpage
\appendix

\section{Pathological Reasoning Templates.}
\label{app:templates}
We present pathological reasoning templates in Tab. \ref{prompts}. System prompt and VQA/ subtyping/ detection prompt are utilized for the task performer, while system prompt and token allocation prompt for the token allocator.

\begin{table}[ht]
\caption{Pathological Reasoning Templates.}
\label{prompts}
\begin{center}
{
\begin{tabular}{p{13.5cm}}  
\toprule
\textbf{System Prompt $\mathsf{p^{sys}}$:} A conversation between User and Assistant. The user asks a question, and the assistant solves it. The assistant first thinks about the reasoning process in the mind and then  provides the user with the answer. The reasoning process and answer are enclosed within <think> </think> and <answer> </answer> tags, respectively, i.e., <think> reasoning process here </think>  <answer> answer here </answer>.\\
\midrule
\textbf{VQA Prompt $\mathsf{p^{vqa}}$:} \emph{\{user\_question\}}?\\
\midrule
\textbf{Subtyping Prompt $\mathsf{p^{sub}}$:} Classify this pathological image into one of these categories: (A) \emph{\{Category\_A\}}, (B) \emph{\{Category\_B\}}, (C) \emph{\{Category\_C\}}...\\
\midrule
\textbf{Detection Prompt $\mathsf{p^{det}}$:} Detect \emph{\{pathological\_category\}} in pathology \emph{\{organ\}}. Output bounding boxes in [[x\_min, y\_min, x\_max, y\_max],...] format.\\
\midrule
\textbf{Token Allocation Prompt $\mathsf{p^{ta}}$:} Allocate the optimal token number for the image based on the pathological task. Generally, simple images and tasks receive fewer tokens and complex ones receive more tokens. The current input token number is \emph{\{current\_token\}} and a maximum limit is \emph{\{max\_token\}}. The pathological task is: \emph{\{VQA /subtyping /detection  prompt\}}. The answer should be a positive integer of the image token number.\\
\bottomrule
\end{tabular}
}
\end{center}
\end{table}

\section{Dataset Details} 
\label{app:datasets}
We evaluate our method on six datasets: PathMMU \cite{sun2024pathmmu}, PathVQA \cite{he2020pathvqa}, UniToPatho \cite{barbano2021unitopatho}, ESCA \cite{tolkach2023artificial}, CRAG \cite{graham2019mild}, and DigestPath2019 Tissue \cite{da2022digestpath}. 
Among these datasets, PathMMU and PathVQA are utilized for VQA, UniToPatho and ESCA are for disease subtyping, while CRAG and and DigestPath2019 Tissue are for lesion detection.

\emph{PathMMU} is a large-scale, multimodal, expert-curated pathology VQA dataset, which is sourced from diverse medical repositories including PubMed scientific articles (PathMed), pathology textbooks (Atlas), educational YouTube videos (EduContent), expert-contributed social media posts (SocialPath), and existing pathology classification datasets (PathCLS). 
Since the training data and the data of the SocialPath subset are not publicly available, we use the official $\operatorname{testtiny}$ and $\operatorname{val}$ data of the other four subsets for evaluation and the rest test data for training. The total amount of training, validation, test data are 6,901, 555, and 921 respectively.

\emph{PathVQA} is a large-scale VQA dataset specifically designed for pathological image analysis, comprising 32,799 open-ended and binary (yes/no) question-answer pairs derived from 4,998 pathological images sourced from publicly available textbooks and the digital library Pathology Education Informational Resource (PEIR). It supports diverse question types, including what, where, how, and yes/no queries, with 50.2\% being open-ended and the rest binary. For training and evaluation, we use the official train-validation-test split (19,654: 6,259: 6,719 QAs).

\emph{UniToPatho} is a subtyping dataset containing 9,536 H\&E stained images extracted from 292 WSIs, designed for colorectal polyp classification and adenoma grading. The dataset includes six diagnostic categories: normal tissue (950 images), hyperplastic polyp (545 images), tubular adenoma with high-grade dysplasia (454 images), tubular adenoma with low-grade dysplasia (3,618 images), tubulo-villous adenoma with high-grade dysplasia (916 images), and tubulo-villous adenoma with low-grade dysplasia (2,186 images). Following standard practice, we use 6,270 and test 2,399 images for training and testing, respectively.

\emph{ESCA} is a subtyping dataset consists of 367,229 images extracted from 320 H\&E-stained whole slide images of esophageal adenocarcinoma and esophagogastric junction adenocarcinoma. The dataset originates from four institutions: University Hospital Cologne (UKK), Landesklinikum Wiener Neustadt (WNS), TCGA, and University Hospital Berlin Charité (CHA). These images are categorized into eleven distinct histological classes: adventitia (71,131 images), lamina propria mucosae (2,173 images), muscularis mucosae (2,951 images), muscularis propria (83,358 images), regression tissue (56,490 images), gastric mucosa (44,416 images), esophageal mucosa (18,561 images), submucosa (22,117 images), submucosal glands (1,516 images), tumor (63,863 images), and ulceration (753 images). Following the standardized split, we use 178,187 images from CHA for training, while 189,142 images from the combined UKK, WNS, and TCGA formed the test set.

\emph{CRAG} dataset contains 213 H\&E colorectal adenocarcinoma image tiles at 20x magnification with full instance-level annotation.  It is originally used for instance segmentation. We divide the slices into 1024$\times$1024 images and generate bounding boxes for detection based on the instance mask, resulting 1,429 and 321 images for training and testing, respectively.

\emph{DigestPath2019 Tissue} dataset consists of total 872 tissue sub-slices from 476 patients with an average size of 5,000$\times$5,000, which are extracted from both benign and malignant areas to cover as much variety of tissue appearance as possible. Like CRAG, it is originally used for instance segmentation. We divide the slices into 1024$\times$1024 images and generate bounding boxes based on the instance mask, resulting 10,725 and 2,666 images for training and testing, respectively.

\begin{figure}[t]
    \centering
    \includegraphics[width=\linewidth]{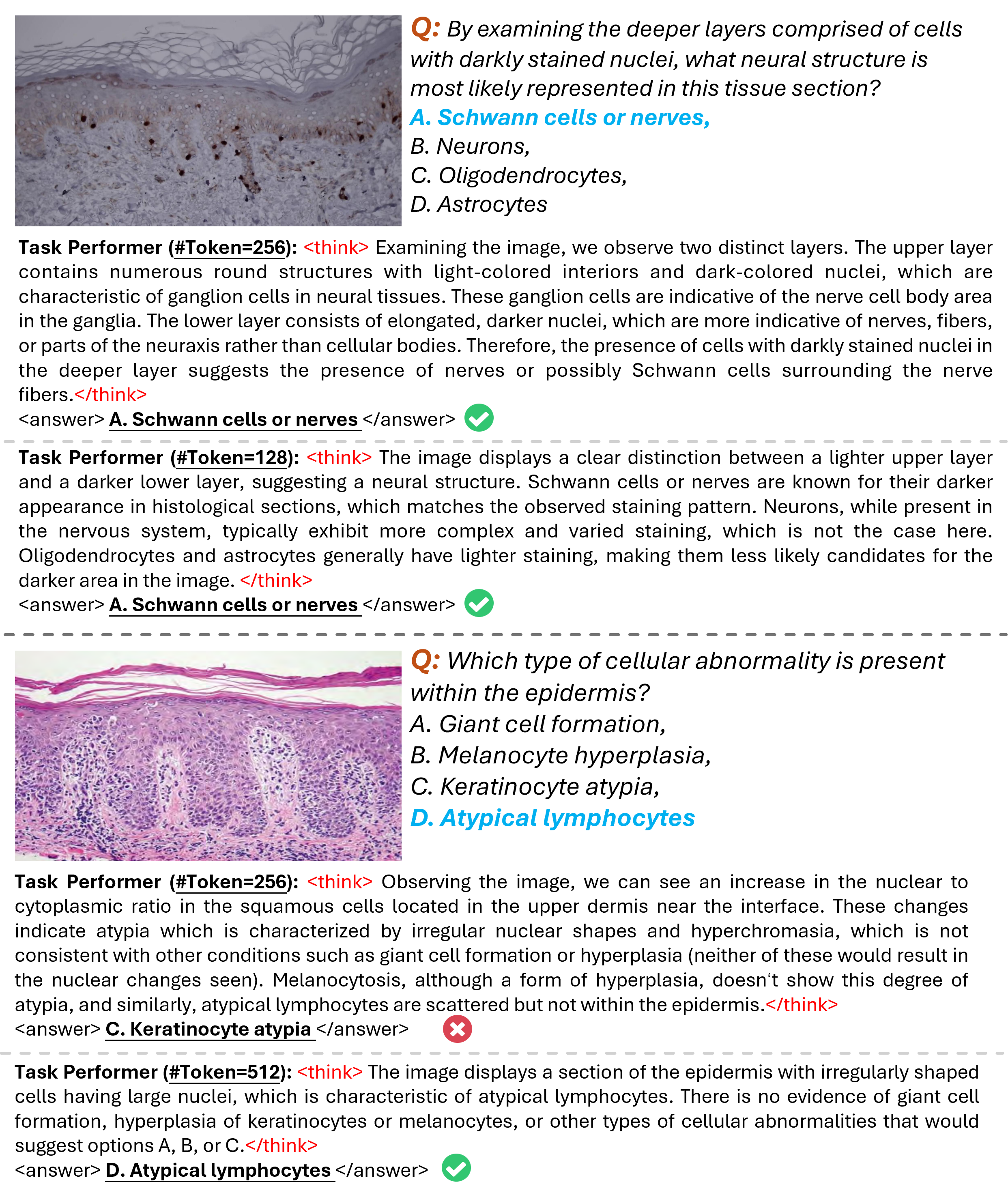}
    \caption{Static token budgets result in suboptimal efficiency-accuracy trade-offs across diverse diagnostic scenarios.}
    \label{fig:motivation}
\end{figure}

\section{Discussion of Token Allocation} 
\label{app:discussion}
While the task performer demonstrates promising pathological reasoning capabilities, we identify a critical limitation in the static token allocation strategy for the pathological images. 
As shown in the first example of Fig. \ref{fig:motivation}, the task performer with both 128 and 256 image tokens correctly identified Schwann cells or nerves (Option A). This suggests that for straightforward cases with well-defined histological features (e.g., ganglion cells with distinct nuclei), a lower token budget (128) suffices for accurate interpretation. Allocating excessive tokens (256) in such scenarios leads to computational over-provisioning without improving diagnostic confidence, thereby wasting resources. 
In the second example (epidermal cellular abnormality), A strict low-token budget (e.g., 256) might fail to capture subtle but critical features (e.g., irregular nuclear contours and increased nuclear-to-cytoplasmic ratio), leading to missed diagnoses of keratinocyte atypia (Option C).

These observations underscore the need for adaptive token allocation, where the model dynamically adjusts its computational budget based on input complexity. Such an approach would allocate minimal tokens for routine cases to preserve computational resources and expand token capacity for challenging cases if necessary.

\begin{table*}[ht]
    \centering
    \caption{The configuration of different pathology-specialized models used for comparison. UDK represents Unified Knowledge Distillation}
    \resizebox{\linewidth}{!}{\begin{tabular}{lcccccc}
        \toprule
         Model&  Data Source& WSIs & Patches & Model arch.  & Model size & Pretraining\\
         \midrule
         Ctranspath & TCGA+PAIP &32K& 4.2M& SwinTrans.&28M&MoCoV3 \\
         Phikon &TCGA&6K& 43M& ViT-B&86M& iBOT  \\
         PLIP & OpenPath & NA & 200K & ViT-B& 86M & CLIP \\
 CHIEF & Public+Private& 60K& 15M& SwinTrans.& 28M&MoCoV3+CLIP\\
 Prov-Gigapath & Private& 171K& 1.3B& ViT-G& 1.1B&DINOv2+MAE\\
         GPFM & 33 Public datasets& 72K & 190M & ViT-L& 307M & UDK \\
         \bottomrule
    \end{tabular}}
    \label{tab:sota_config}
\end{table*}

\section{Additional Implementation Details}
\label{app:implementation}
For the cancer subtyping task, we compare our method with traditional pathology-specialized models (\emph{i.e.}, 
Ctranspath \cite{wang2022transformer}, Phikon \cite{filiot2023scaling}, PLIP \cite{huang2023visual}, CHIEF \cite{wang2024pathology}, Prov-Gigapath \cite{xu2024whole}, and GPFM \cite{ma2024towards}) with pathology-specific pre-training. Tab.\ref{tab:sota_config} shows the configuration of different models.  We obtained the pre-trained models from official sources and followed the original papers to train ABMIL \cite{ilse2018attention} with a learning rate of 2e-4 for 30 epochs.

\section{Additional Quantitative Results}

We further compare our method with BiomedParse \cite{zhao2025foundation} for the detection task. Specifically, BiomedParse is a biomedical foundation model that can jointly conduct segmentation, detection, and recognition across nine imaging modalities including pathology, which is trained using 6 million biomed images (15.5K pathology images).
Both of the zero-shot and fine-tuned results are presented, which are obtained following its official setup. The learning rate and iterative epoch for fine-tuning are set as 1e-5 and 20, respectively. As shown in Tab. \ref{tab:dg19tissue_supp} and \ref{tab:crag_supp}, the zero-shot version performs poorly, indicating its inability to adapt to the new dataset without task-specific training. The fine-tuned BiomedParse shows significant improvement but remains substantially inferior to subsequent methods, suggesting its limitation for pathology lesion detection. Our proposed framework achieves state-of-the-art detection performance on the DigestPath2019 Tissue and CRAG datasets while maintaining computational efficiency.

\begin{table}
\centering{
\caption{Detection performance of different models on the DigestPath2019 Tissue dataset. }\label{tab:dg19tissue_supp}%
    % \resizebox{\linewidth}{!}{
\footnotesize
\begin{tabular}{l c c c c c c }
\toprule
Method &mAP@0.1 &mAP@0.3 &mAP@0.5 &AVG &TPI &Reasoning\\
\midrule
BiomedParse (Zero-Shot) &17.2&14.2&11.5&14.3 &-&\ding{55} \\
BiomedParse (Fine-Tuned) &35.5&32.1&27.2& 31.6&-&\ding{55} \\
\midrule
Qwen2.5-VL-7B &22.6 &7.1 &2.2 &10.6 &1369.0&\ding{55} \\
SFT &59.3 &54.1 &46.9 &53.4 &\underline{256.0} &\ding{55}\\
\rowcolor{gray!15}+Performer (Ours) &\underline{77.1} &\underline{71.0} &\underline{61.6} &\underline{69.9} &\underline{256.0} &\ding{52}\\
\rowcolor{gray!15}+Performer+Allocator (Ours)&\textbf{82.1}&\textbf{74.7}&\textbf{63.0}&\textbf{73.3}&\textbf{244.0}&\ding{52}\\
\bottomrule
\end{tabular}
}
% }
\end{table}

\begin{table}
\centering{
\caption{Detection performance of different models on the CRAG dataset. }\label{tab:crag_supp}%
    % \resizebox{\linewidth}{!}{
\footnotesize
\begin{tabular}{l c c c c c c}
\toprule
Method &mAP@0.1 &mAP@0.3 &mAP@0.5 &AVG &TPI &Reasoning\\
\midrule
BiomedParse (Zero-Shot) &7.9&6.6&4.6&6.4 &-&\ding{55} \\
BiomedParse (Fine-Tuned) &37.5&35.1&30.9&34.5 &-&\ding{55} \\
\midrule
Qwen2.5-VL-7B &24.8 &7.6 &1.9 &11.4 &1369.0 &\ding{55}\\
SFT &62.3 &59.2 &54.2 &58.6 &\underline{256.0} &\ding{55}\\
\rowcolor{gray!15}+Performer (Ours) &\underline{77.0} &\underline{73.3} &\textbf{66.0} &\underline{72.1} &\underline{256.0} &\ding{52}\\
\rowcolor{gray!15}+Performer+Allocator (Ours)&\textbf{81.7}&\textbf{75.6}&\underline{64.5}&\textbf{73.9}&\textbf{229.8} &\ding{52}\\
\bottomrule
\end{tabular}
}
% }
\end{table}

\section{Additional Qualitative Results} 
\label{app:qualitative}

\begin{figure}[ht]
    \centering
    \includegraphics[width=\linewidth]{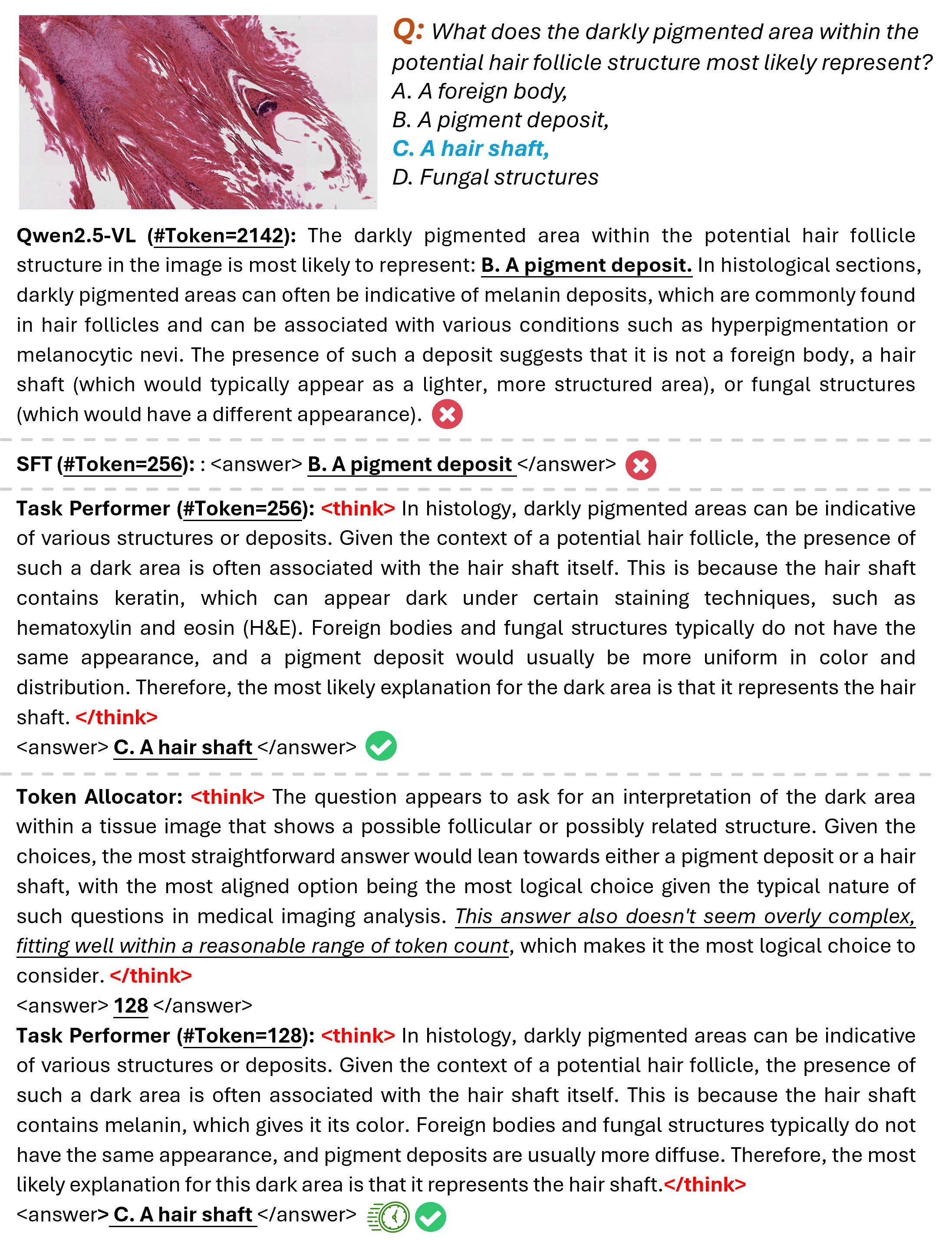}
    \caption{Diagnostic divergence in interpreting pigmented hair follicle structures. Qwen2.5-VL (2142 image tokens) and SFT misclassifies the dark area as "pigment deposit", whereas task performer (256 tokens) identifies the correct "hair shaft" by referencing keratin/melanin properties under H\&E staining. The token allocator’s reduction to 128 tokens highlights computational efficiency without compromising diagnostic fidelity. }
    \label{fig:qua2}
\end{figure}

\begin{figure}[ht]
    \centering
    \includegraphics[width=\linewidth]{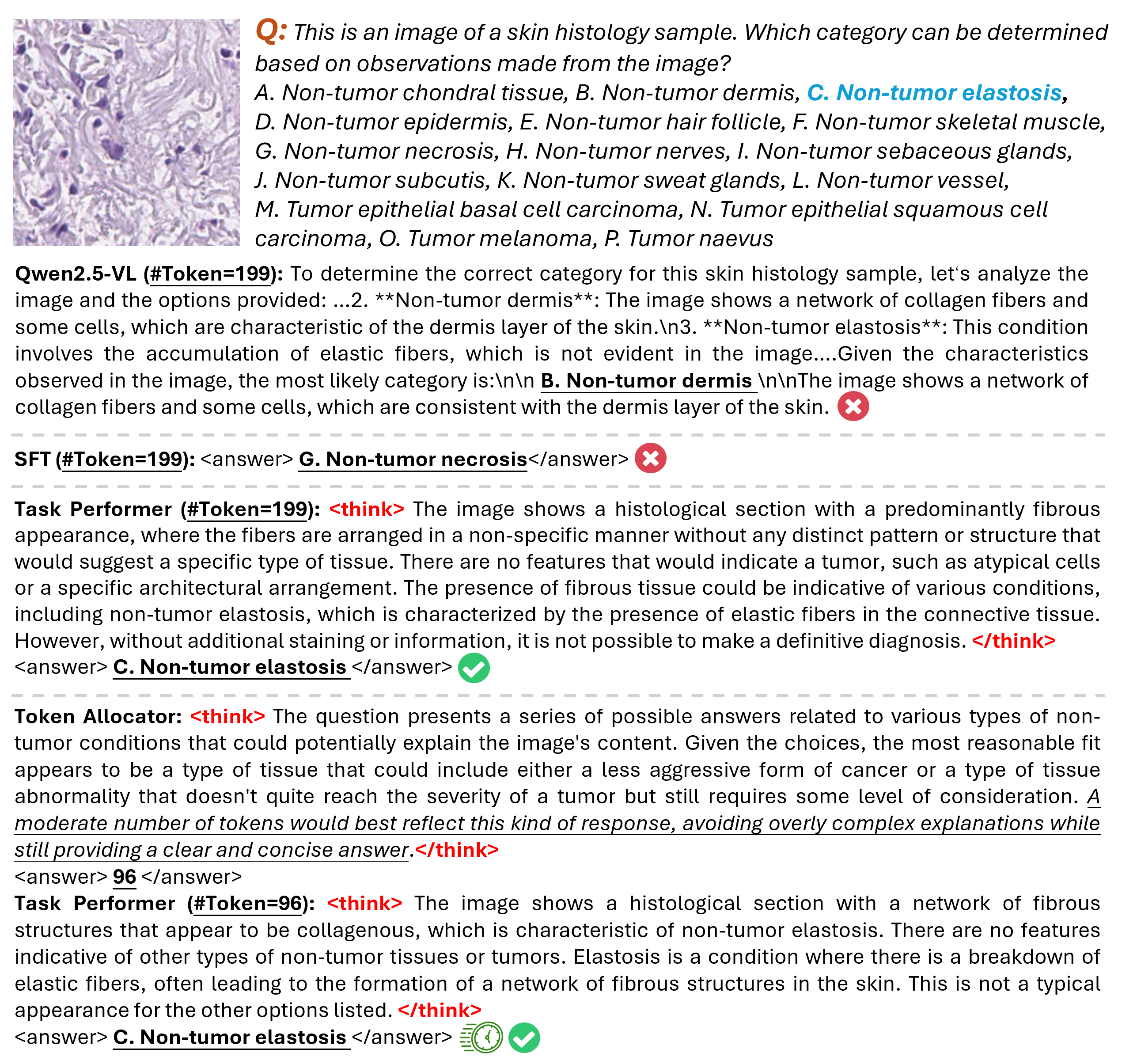}
    \caption{Performance comparison of Qwen2.5-VL, SFT, and task performer models in classifying a non-tumor skin histology sample. While Qwen2.5-VL incorrectly identifies the tissue as "non-tumor dermis", and SFT provides an unsupported diagnosis of "non-tumor necrosis", our task performer correctly classifies it as "non-tumor elastosis" with interpretable reasoning about fibrous structures. The token allocator further optimizes efficiency by reducing tokens to 96 without sacrificing accuracy.}
    \label{fig:qua3}
\end{figure}

\begin{figure}[ht]
    \centering
    \includegraphics[width=\linewidth]{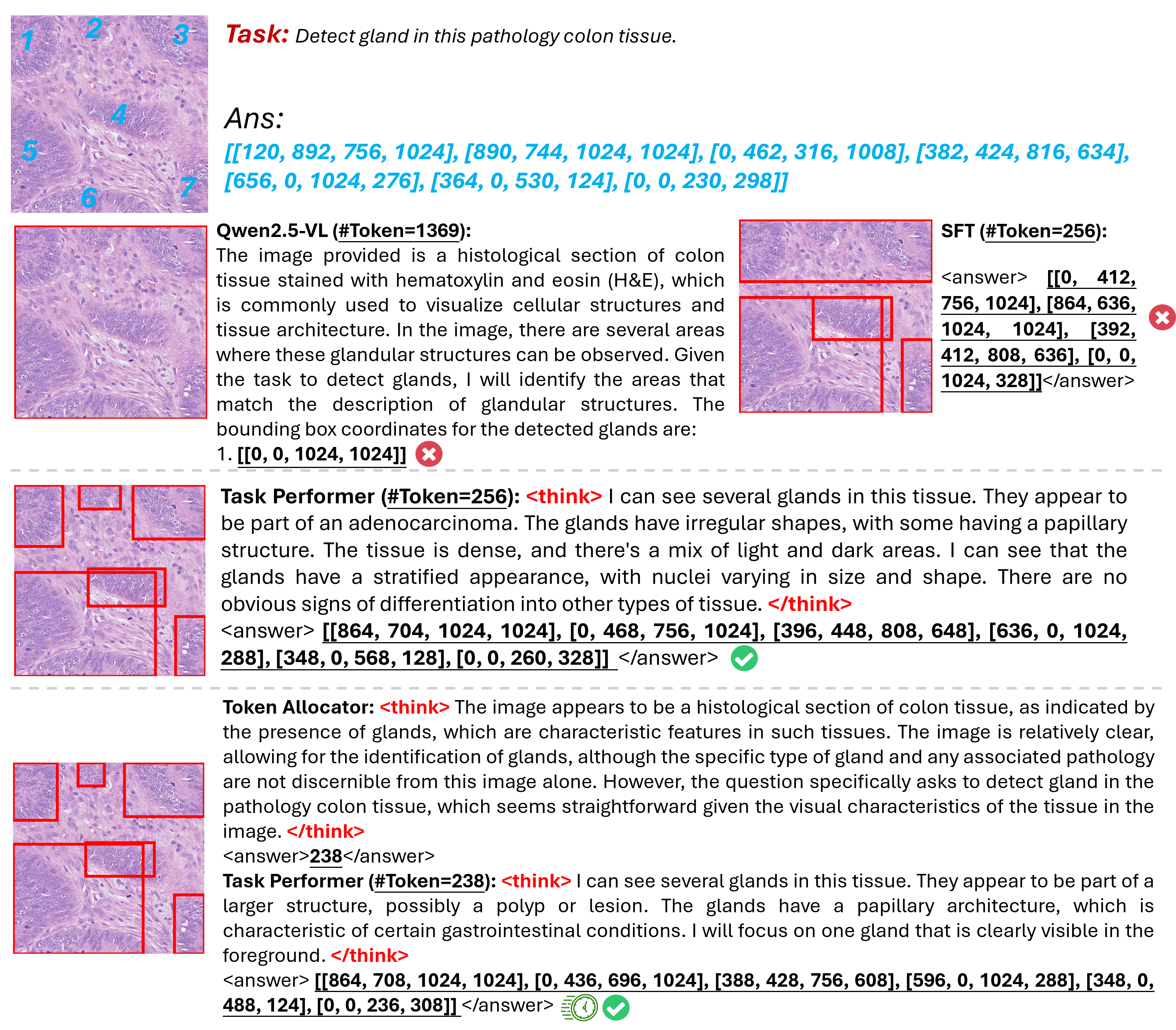}
    \caption{Detection of glandular structures in colon tissue across models. Qwen2.5-VL fails to localize glands precisely (outputting a full-image bounding box), while SFT generates partially accurate coordinates. Our task performer (256 tokens) identifies irregular glandular architectures suggestive of adenocarcinoma, and its 238-token variant maintains precision with refined bounding boxes (e.g., [864, 708, 1024, 1024]).}
    \label{fig:qua4}
\end{figure}

We present additional qualitative results in Fig. \ref{fig:qua2}, \ref{fig:qua3} and \ref{fig:qua4}. The evaluation of three distinct histopathological tasks reveals consistent patterns in model capabilities and limitations. General-purpose Qwen2.5-VL model exhibit critical shortcomings in pathological image analysis, as demonstrated by their misclassification of non-tumor dermis (B) instead of elastosis (C) in skin histology, and their oversimplified bounding box prediction ([[0, 0, 1024, 1024]]) for gland detection, which lacks granularity. These errors stem from insufficient domain-specific knowledge and an inability to discern fine-grained pathological features. Supervised fine-tuned (SFT) models, while using less tokens, compromise diagnostic transparency, as seen in their terse outputs (e.g., "Non-tumor necrosis" without justification) and inconsistent gland localization (e.g., incomplete bounding boxes).

In contrast, the task performer framework addresses these gaps by integrating domain-adapted reasoning. For skin histology, it correctly identifies non-tumor elastosis (C) through structured analysis of fibrous tissue patterns. In gland detection, it outperforms baselines by generating precise bounding boxes (e.g., [[864, 708, 1024, 1024]]) supported by contextual reasoning about glandular architecture, even with reduced tokens (238 vs. 256). Similarly, for hair follicle interpretation, it consistently recognizes the hair shaft (C) by leveraging histological knowledge (e.g., keratin/melanin properties), whereas Qwen2.5-VL erroneously attributes the pigmentation to deposits (B).

The token allocator further enhances efficiency, dynamically adjusting budgets (e.g., 128 tokens for hair follicle analysis) to match task complexity. This adaptability prevents over-provisioning for straightforward tasks (e.g., pigment vs. hair shaft discrimination) while reserving computational resources for nuanced challenges like gland detection. Collectively, these results underscore that specialized reasoning and adaptive computation are pivotal for robust pathology AI, bridging the accuracy-interpretability-efficiency trade-off that plagues generalist models and static approaches.

\end{document}